\documentclass{article}

\PassOptionsToPackage{numbers, compress}{natbib}

   \usepackage[preprint]{neurips_2020}

\usepackage[utf8]{inputenc} 
\usepackage[T1]{fontenc}    
\usepackage{hyperref}       
\usepackage{url}            
\usepackage{booktabs}       
\usepackage{amsfonts}       
\usepackage{nicefrac}       
\usepackage{microtype}      
\usepackage{wrapfig}
\usepackage{graphicx}
\usepackage{amsmath}
\usepackage{multirow}

\usepackage[labelformat=simple]{subcaption}

\usepackage{color}

\title{ADMP: An Adversarial Double Masks Based Pruning Framework For Unsupervised Cross-Domain Compression}

\author{%
  Xiaoyu Feng\\
  Department of Electronic Engineering\\
  Tsinghua University\\
  \texttt{feng-xy18@mails.tsinghua.edu.cn} \\
  \And
  Zhuqing  Yuan\\
  Department of Electronic Engineering\\
  Tsinghua University\\
  \texttt{yuanzhuqing@tsinghua.edu.cn} \\
  \AND
  Guijin Wang \\
  Department of Electronic Engineering\\
  Tsinghua University\\
  \texttt{wangguijin@tsinghua.edu.cn} \\
  \And
  Yongpan Liu \\
  Department of Electronic Engineering\\
  Tsinghua University\\
  \texttt{ypliu@tsinghua.edu.cn} \\
}

\begin{document}

\maketitle

\begin{abstract}
 Despite the recent progress of network pruning,  directly applying it to the Internet of Things (IoT) applications still faces two challenges, \textit{i.e.} the distribution divergence between end and cloud data and the missing of data label on end devices. One straightforward solution is to combine the \textit{unsupervised domain adaptation} (UDA) technique and pruning. For example, the model is first pruned on the cloud and then transferred from cloud to end by UDA. However, such a naive combination faces high performance degradation. Hence this work proposes an Adversarial Double Masks based Pruning (ADMP) for such cross-domain compression. In ADMP, we construct a Knowledge Distillation framework not only to produce pseudo labels but also to provide a measurement of domain divergence as the output difference between the full-size teacher and the pruned student. Unlike existing mask-based pruning works, two adversarial masks, \textit{i.e.} soft and hard masks, are adopted in ADMP. So ADMP can prune the model effectively while still allowing the model to extract strong domain-invariant features and robust classification boundaries. During training, the Alternating Direction Multiplier Method is used to overcome the binary constraint of $\{ 0,1 \}$-masks. On Office31 and ImageCLEF-DA datasets, the proposed ADMP can prune $60\%$ channels with only $0.2\%$ and $0.3\%$ average accuracy loss respectively. Compared with the state of art, we can achieve about $1.63\times$ parameters reduction and $4.1\%$ and $5.1\%$ accuracy improvement.
\end{abstract}

\section{Introduction}
Nowadays, one mature solution to apply neural networks to IoT devices, such as drones or portable wireless electrocardiogram monitors, is to use the end-cloud system where the task of model training is offloaded to the cloud while the end device only focuses on model inference. But most neural networks are memory- and computation-intensive compared with the limited resources of IoT end devices. Hence model compression techniques like \textit{pruning} \cite{alvarez2016learning, he2017channel, luo2017thinet, yu2018nisp, zhuang2018discrimination, you2019gate} are often needed. With the help of pruning, the size of the model trained on the cloud can be effectively reduced while retaining the performance. 

Although the deep learning community has witnessed the success of pruning these years, there are still many challenges in applying pruning to the end-cloud IoT systems. In most IoT applications, there is significant distribution divergence between the real data collected by sensors on end devices and the on-cloud datasets used for model training. Such data divergence may cause severe model performance degradation. Even more difficult is that end data are unlabeled in most cases. This makes it hard to use traditional supervised learning based pruning. 
People familiar with \textit{unsupervised domain adaptation} (UDA), which is effective for transferring model from a supervised domain to an unsupervised domain, may argue that such problems can be solved by combing the two techniques together. For example, the model can be first pruned on source domain (cloud) and then adapted to the target domain (end). However, we must point out that \cite{yu2019accelerating, chen2019cooperative} have shown that such a simple combination has poor performance in practice and the cross-domain pruning remains a challenge.
 
In this work, we propose a new \textit{Adversarial Double Masks based Pruning} framework (ADMP) for cross-domain compression. In ADMP, the full-size baseline and the pruned model form a Knowledge Distillation (KD) \cite{hinton2015distilling} framework. Based on it, we propose a method to measure the domain discrepancy during pruning. Although formulating and decreasing domain divergence have always been the main tasks of UDA \cite{ben2010theory, ben2007analysis,long2015learning,long2016unsupervised}, how to formulate domain discrepancy during pruning is 
rarely studied. So to simplify the search space and be more in line with the process of pruning, we construct domain discrepancy as the upper bound of the output difference between the pruned student and full-size teacher on the target domain.

In recent years, more and more researchers are focusing on the now popular adversarial UDA methods \cite{tzeng2017adversarial,saito2018maximum,zhang2019domain,lee2019drop}. Compared with traditional methods, they can better align the two domains while moving the decision boundaries away from the sample-dense area in feature space. Motivated by such view, ADMP also contains two adversarial stage, \textit{channel search} and \textit{adversarial update}. Unlike traditional pruning works, two adversarial masks, \textit{i.e.} soft and hard masks, are adopted in ADMP. In \textit{channel search} stage, the hard mask searches sub-structures that have most different output from the teacher. While the network weights and soft mask are optimized adversarially to minimize this performance difference in \textit{adversarial update} stage. Through such adversarial optimization process, the domain divergence can be effectively minimized during pruning. 

Our contributions can be summarized as follows: (1) We propose a KD based measurement of domain discrepancy which can be in good line with the process of pruning. (2) We propose an adversarial two-stage pruning algorithm. Using two adversarial pruning masks, the domain discrepancy can be effectively minimized during pruning. (3) Experiments show that ADMP outperforms the state of art. On Office31 and ImageCLEF-DA, ADMP can achieve $1.63\times$ parameters reduction and $4.1\%$ and $5.1\%$ accuracy improvement. We also find that ADMP performs much better for the difficult transfer tasks, e.g. $D \rightarrow A$ on Office31.
\section{Related works}
\label{related}
\textbf{\textit{Unsupervised domain adaptation}} tackles the problem that no data labels are available on the target domain. According to Ben-David \textit{et al.} \cite{ben2010theory,ben2007analysis}, error on the target domain can be optimized by minimizing error on the source domain and domain discrepancy. Long \textit{et al.} \cite{long2015learning,long2016unsupervised} introduce Maximum Mean Discrepancy (MMD) into the loss function to make feature extracted on two domains similar. Apart from using pre-defined distance metrics like MMD, Ganin \textit{et al.} \cite{ganin2016domain} propose a GAN-like structure so the network can automatically learn how to extract features that are domain invariant. Motivated by their work, the adversarial UDA methods become very popular. MCD \cite{saito2018maximum} uses two different classifiers to find the samples that are more likely to be misclassified. Lee \textit{et al.} \cite{lee2019drop} use adversarial dropout to produce the classification inconsistency required by $\mathcal{H}\Delta \mathcal{H}$-distance.
Zhang \textit{et al.} \cite{zhang2019domain} extend \cite{ganin2016domain} by integrating category information in the discriminator so the domain can be aligned at a finer granularity. In this work, domain discrepancy is measured by output difference between the full-size teacher and pruned student on the target domain. An adversarial two-stage algorithm is proposed to minimize it and prune the model at the same time.

\textbf{\textit{Pruning}} is an effective method for compressing and accelerating neural networks. LeCun \textit{et al.} \cite{lecun1990optimal} first use second-derivative to prune the neural network while Han \textit{et al.} \cite{han2015learning} choose the absolute value to speed up pruning. Given that the irregular sparsity pattern of their methods requires dedicated hardware to achieve acceleration, many researchers have turned their attention to structured pruning. Wen \textit{et al.} \cite{wen2016learning} use group Lasso to decide which filters should be pruned. Further, He \textit{et al.} \cite{he2017channel} and Luo \textit{et al.}  \cite{luo2017thinet}  use the influence of each pruned filter on following layers as the indicator. Instead of using the norm-induced criterion, He \textit{et al.} \cite{he2019filter} point out the weights near the geometric median in a layer are less useful. Zhuang \textit{et al.} \cite{zhang2019domain} use an attention mechanism to train the pruned network to be more informative. In this work, we mainly focus on structured pruning. 

\textbf{\textit{Cross-domain pruning}} Chen \textit{et al.} \cite{chen2019cooperative} focus on unstructured cross-domain pruning and propose a dynamic and cooperative pruning method. Yu \textit{et al.} \cite{yu2019accelerating} propose a Taylor-based strategy to search the filters to be pruned and they introduce MMD into their optimization objective to reduce domain discrepancy. However, their methods face severe performance degradation after pruning. So this task remains a challenge. The proposed ADMP can better integrate the philosophy of UDA into the process of pruning so ADMP can achieve higher accuracy and sparsity after pruning. 


\section{Methodology}
\label{method}
\subsection{Overall idea and formulation}
\label{sec:idea}
\begin{figure}
    \centering
    \includegraphics[width=0.9\textwidth]{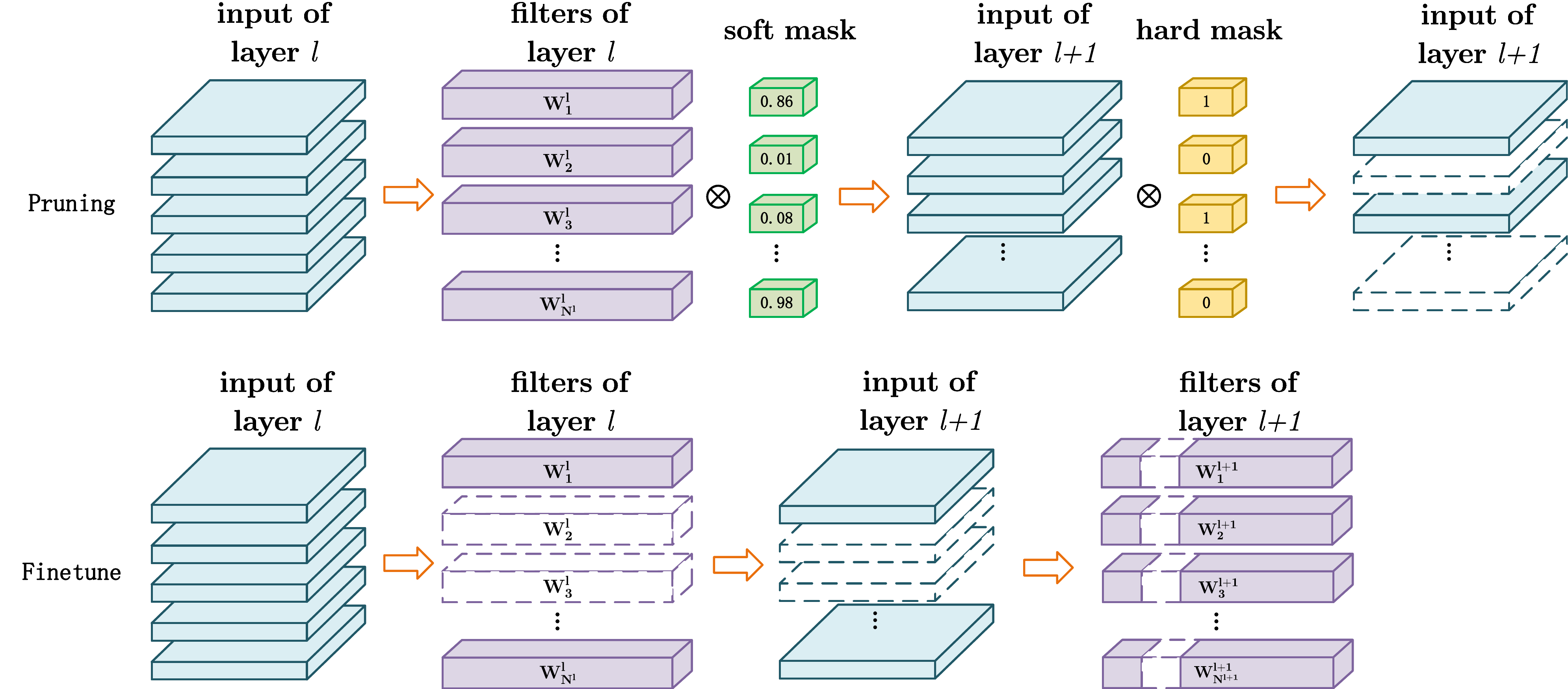}
    \caption{(Best viewed in color) An illustration of the process of pruning and fine-tuning. The dashed blocks represent the filters and feature maps that are pruned. In pruning, the soft mask only scales the weights so all weight filters are kept. While the hard mask prunes corresponding filters of the output activation. But in fine-tuning, the hard mask is dropped before training. Then the filters of layer $l$ and input channels of layer $l+1$ are pruned based on the soft mask of layer $l$.}
    \label{fig:prune+finetune}
\end{figure}
Before giving the problem formulation, it is necessary to first introduce UDA in general. In UDA problems, both data $X_s$ and labels $Y_s$ are available on the source domain but only the data $X_t$ on the target domain can be accessed. The optimization objective is to minimize the error on the target domain $\mathcal{R}_T$. According to the theories proposed by Ben-David \textit{et al.} \cite{ben2007analysis,ben2010theory}, $\mathcal{R}_T$ is bounded by source error $\mathcal{R}_S$ and the domain discrepancy. In most cases, $\mathcal{R}_S$ can be denoted as the softmax entropy $L_{\mathcal{S}}(X_s, Y_s)$ and easily optimized in a supervised way. So the core is how to minimize the domain discrepancy. Unlike early works that use pre-defined metrics for domain discrepancy, the adversarial methods have become one popular trend\cite{ganin2016domain,saito2018maximum,lee2019drop,zhang2019domain}. The reason is that classifiers trained by traditional methods often have bad decision boundaries that cross the sample-dense area in the feature space. However, by using an adversarial training process, the network can be pushed to learn how to effectively separate the decision boundaries and the sample area. Hence, the remaining question is how to integrate adversarial UDA into pruning.

According to the theory of Ben-David \textit{et al.}\cite{ben2010theory}, the key to decreasing domain discrepancy is to find the maximal output difference between any two classifiers and minimize it. On the other hand, we can consider pruning as a two-step process, searching for one sub-structure, and updating the model parameters. Hence our idea is to propose an adversarial two-stage pruning method. The first stage aims to explore different sub-structures to find the largest output difference. Then the network parameters are updated in the adversarial direction to minimize such output difference. However, it is hard to explore all possible sub-structures. So we propose a KD based framework where one classifier is fixed to the full-size network $\mathcal{A}_0$. On the other hand, the teacher output can be used as pseudo labels. So the problem of missing labels can be relieved. In this way, the search stage aims to find one sub-structure $\mathcal{A} \in \mathbb{A}$ that differs most from $\mathcal{A}_0$ where $\mathbb{A}$ is space spanned by all sub-structures of $\mathcal{A}_0$. We denote the parameters of $\mathcal{A}$ as $\mathcal{W}_{\mathcal{A}}$. So we can give the following optimization objectives: 
\begin{equation}
\label{eq1}
 \underset{\mathcal{W}_{\mathcal{A}}}{min} \quad L_{\mathcal{S}}(X_s,Y_s;\mathcal{A}) + |\underset{x \sim \mathcal{X}_s}{\bf{E}}I[\mathcal{A}(x) \neq \mathcal{A}_0(x)] - \underset{x \sim \mathcal{X}_t}{\bf{E}}I[\mathcal{A}(x) \neq \mathcal{A}_0(x)]|
\end{equation}
where
\begin{equation}
\label{eq2}
    \mathcal{A} = \underset{\mathcal{A'} \in \mathbb{A}}{argmax}|\underset{x \sim \mathcal{X}_s}{\bf{E}}I[\mathcal{A'}(x) \neq \mathcal{A}_0(x)] - \underset{x \sim \mathcal{X}_t}{\bf{E}}I[\mathcal{A'}(x) \neq \mathcal{A}_0(x)]|
\end{equation}
Here $I[ \cdot ]$ represents the indicator function, hence $\underset{x}{\bf{E}}I[\mathcal{A}(x) \neq \mathcal{A}_0(x)]$ represents the output difference between $\mathcal{A}$ and $\mathcal{A}_0$. We assume $\underset{x \sim \mathcal{X}_s}{\bf{E}}I[\mathcal{A}(x) \neq \mathcal{A}_0(x)] \approx 0$ because source domain is supervised. Then we use $L1$ norm loss to replace the non-differential indicator function on target domain. Now the whole optimization problem can be re-formulated as:
\begin{equation}
\label{eq3}
\begin{aligned}
   & \underset{\mathcal{W}_{\mathcal{A}}}{min} \quad L_{\mathcal{S}}(X_s,Y_s;\mathcal{A}) + \underset{x \sim \mathcal{X}_t}{\bf{E}}\frac{1}{K}\sum_{k=1}^K |p_{\mathcal{A'} k}(x) - p_{\mathcal{A}_0 k}(x)|\\
   & object \  to \quad \mathcal{A} = \underset{\mathcal{A'} \in \mathbb{A}}{argmax}\underset{x \sim \mathcal{X}_t}{\bf{E}}\frac{1}{K}\sum_{k=1}^K |p_{\mathcal{A'} k}(x) - p_{\mathcal{A}_0 k}(x)| \quad card(\mathcal{A}) \leq t
\end{aligned}
\end{equation}
Here $t$ denotes pruning sparsity threshold, $K$ is the number of categories and $p$ denotes the output softmax probability.
\subsection{Pruning framework}
\label{pruning}
\begin{figure}
  \begin{subfigure}{.48\textwidth}
  \centering
  \includegraphics[width=6.5cm]{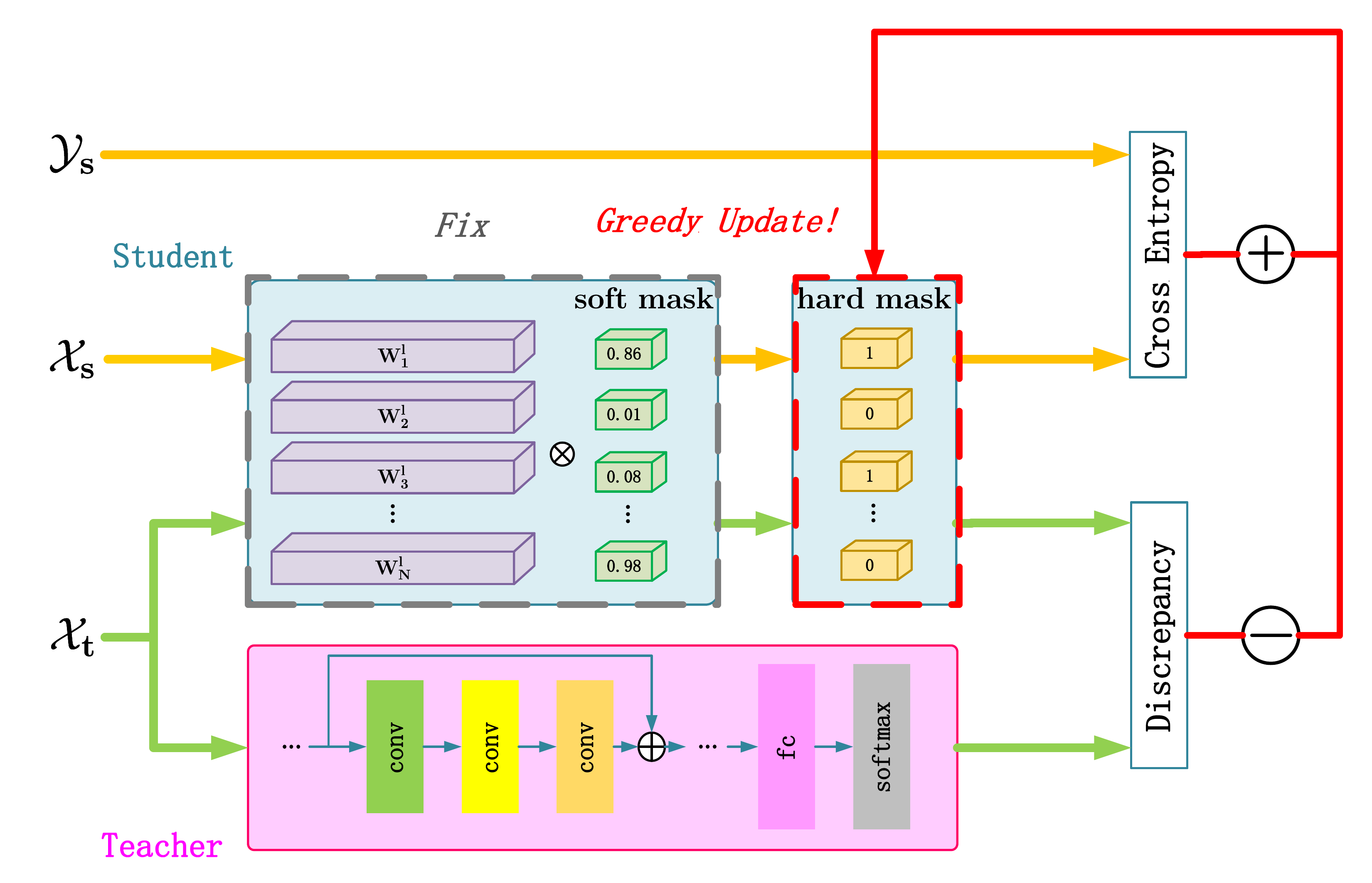}
  \caption{Channel search}
  \label{f-a}
  \end{subfigure}
  \centering
  \begin{subfigure}{.48\textwidth}
  \centering
  \includegraphics[width=6.5cm]{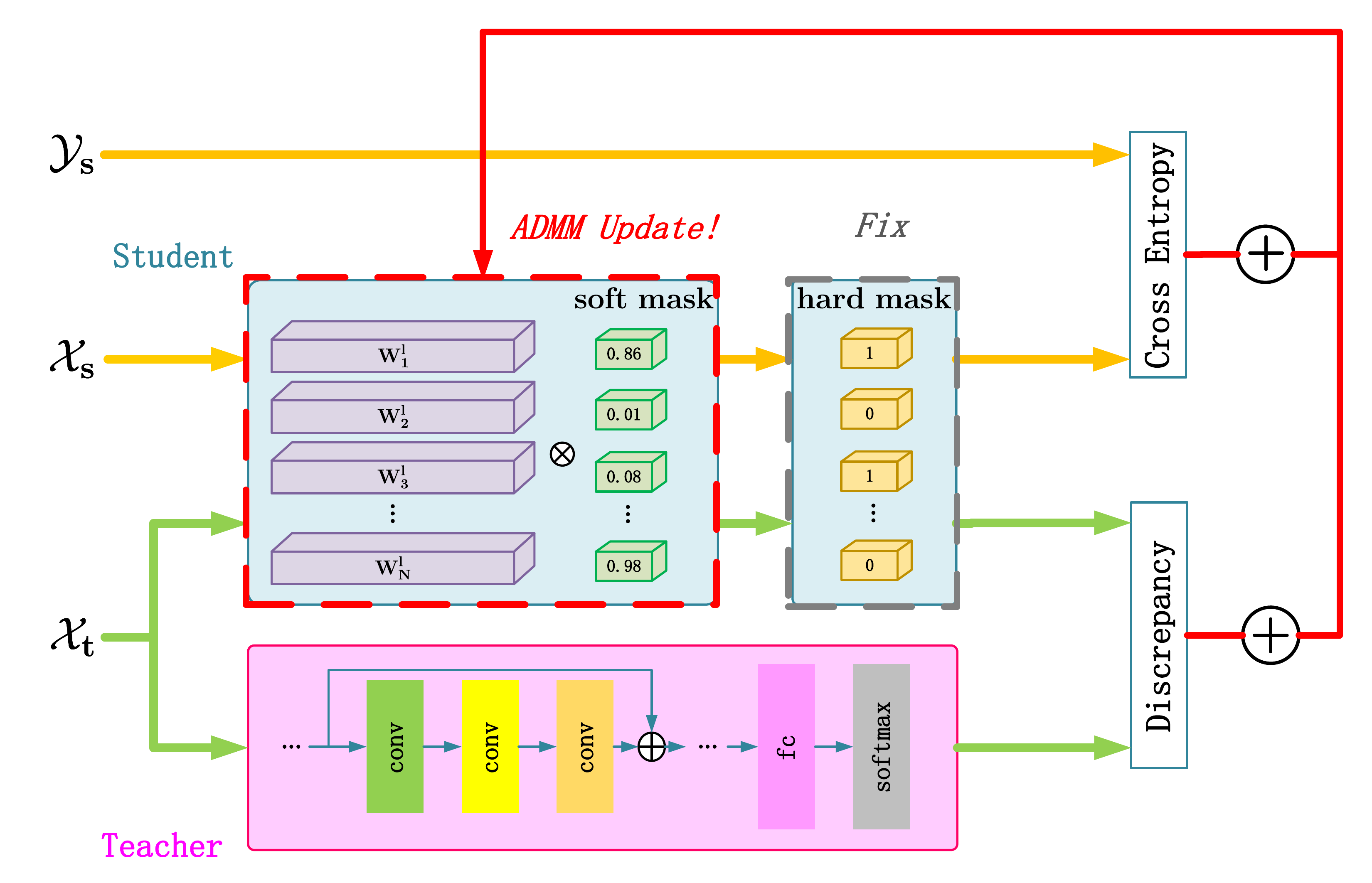}
  \caption{Adversarial update}
  \label{f-b}
  \end{subfigure}
  \caption{(Best viewed in color) The training process of ADMP. In \textit{channel search} stage, $m_h$ is generated by a greedy algorithm based on fixed $W$ and $m_s$. It is optimized to minimize source error but maximize the output difference between it and the teacher on the target domain. In \textit{adversarial update} stage, $m_s$ and $W$ are optimized to minimize both source error and the output difference. The different optimization direction between $m_h$ and $W$ and $m_s$ constructs an adversarial process and it pushes $m_s$ and $W$ to be more robust to the domain divergence.}
  \label{fig:framework}
\end{figure}
 To solve Eq. \ref{eq3}, the whole algorithm mainly consists of three steps: (1)\textit{\textbf{Pretraining}} A full-size model is first trained by a standard UDA process. (2)\textit{\textbf{ADMP}} is the core of whole algorithm. To prune the network, we follow the common way in structured pruning to use one $\{ 0,1 \}$-mask to represent the pruned filters of each layer. Now, the search of $\mathcal{A}$ in Eq. \ref{eq3} can be transferred to the learning of mask ${\{ m^l \}}_{l=1}^L$. However, we introduce two adversarial masks, \textit{i.e} soft and hard masks ($m_s$ and $m_h$), in this work. As shown in Fig. \ref{fig:prune+finetune}, $m_h$ is strictly binary while $m_s$ has continuous value. The reasons for using two masks are that we hope to find the $\mathcal{A}$ in Eq.\ref{eq3} to expose the domain discrepancy. But its optimization direction indicates it is not a good structure for the pruned model. So during pruning, $m_h$ maximizes the output difference between the student and teacher on the target domain to search $\mathcal{A}$. While $W$ and $m_s$ take the feedback of $m_h$ and are optimized in an adversarial direction to minimize both source error and domain discrepancy. Although $m_s$ represents the final searched structure, it only scales the weights but not prunes them during pruning. (3)In \textit{\textbf{Fine-tuning}}, $m_h$ is dropped. Then the filters with corresponding $m_s$ values close to zero are pruned. Such a process makes sure that the domain discrepancy is effectively exposed while the final structure still has good performance. After pruning output filters in $l$-layer, input channels of $l+1$-layer are also pruned according to the $m_s^l$ in $l$-layer as shown in Fig. \ref{fig:prune+finetune}. In the following subsections, we will further explain the training details of ADMP.
\subsubsection{Channel search} 
There are two stages in ADMP, \textit{channel search} and \textit{adversarial update}. Given that the Eq. \ref{eq3} is not easy to solve, we further split the original problem into two alternating sub-problems to separate $W$, $m_s$ and $m_h$. Fig. \ref{fig:framework} illustrates the alternating updating process of the two stages. In \textit{channel search} stage, as Fig. \ref{f-a} shows, both the model parameters and soft mask are fixed and only the hard mask is updated. According to Eq. \ref{eq3}, we present the optimization objective of \textit{channel search} as:
\begin{equation}
\label{eq4}
    \underset{m_h}{min} \quad \mathcal{L}_s(X_s,Y_s) \  - \underset{x \sim \mathcal{X}_t}{\bf{E}}\frac{1}{K}\sum_{k=1}^K |p(x;W\odot m_s,m_h) - p(x;W)|
\end{equation}
Here we denote $p_{\mathcal{A}}(x)=p(x;W \odot m_s, m_h)$ and $p_{\mathcal{A}_0}(x)=p(x;W)$ to represent the softmax output of the pruned and unpruned model. The first item is to make sure the pruned model has accurate classification on the source domain so the condition $\underset{x \sim \mathcal{X}_s}{\bf{E}}I[\mathcal{A}(x) \neq \mathcal{A}_0(x)] \approx 0$ can be met. While the minus sign in the second part means $m_h$ is optimized to maximize the output difference between the student and teacher. Unlike the common masks in traditional pruning, $m_h$ is a temporary mask which means it is regenerated at each iteration. So for quick optimization speed, $m_h$ is updated by a greedy algorithm. Based on $m_s$ and $W$ at each iteration, the importance of filters is quickly evaluated by their contribution to minimizing Eq. \ref{eq4}. 
In this work, we use first-order Taylor expansion to measure the contribution. Then filters with small importance are pruned. 
\subsubsection{Adversarial update}
Unlike $m_h$, the soft mask $m_s$ decides the final structure of the pruned model. In this work, we use the ADMM algorithm in \cite{wu2018ell,li2019compressing} so $m_s$ can be updated together with the weights by the gradient descent. The details of ADMM will be discussed later. As shown in Fig. \ref{f-b}, the hard mask is fixed while both $m_s$ and $W$ are optimized in the direction against it on the target domain. It means both source error and domain discrepancy are minimized as Eq. \ref{eq5} shows. 
\begin{equation}
\label{eq5}
    \underset{W, m_s}{min} \quad \mathcal{L}_s(X_s,Y_s) \  + \underset{x \sim \mathcal{X}_t}{\bf{E}}\frac{1}{K}\sum_{k=1}^K |p(x;W\odot m_s, m_h) - p(x;W)|
\end{equation}
The pruned model tend to have worse performance than the full-size baseline. Given that $m_h$ pushes the pruned model far away from the teacher on the target domain, it may bring a potential negative influence on the model performance. In order to overcome such potential degradation, we use a stronger alignment during \textit{adversarial update}. To be specific, we introduce a clustering loss \cite{luo2018smooth} to Eq. \ref{eq5} to achieve semantic-level data alignment on the target domain. 
\begin{equation}
\label{eq6}
    \frac{1}{|\mathcal{X}_t|^2}\sum_{i=1}^{|\mathcal{X}_t|}\sum_{j=1}^{|\mathcal{X}_t|}\delta_{ij}d(x_i,x_j) + (1 - \delta_{ij})max(0,c - d(x_i,x_j))
\end{equation}
where $\delta_{ij}$ indicates the category predictions of the teacher network. It equals to $1$ when the teacher judges $x_i$ and $x_j$ come from the same category, otherwise it equals to $0$. $d(x_i,x_j)=||p_{\text{student}}(x_i)-p_{\text{student}}(x_j)||_2$ represents the Euclidean distance between student outputs. $c$ controls the distance among different classes. Eq. \ref{eq6} pushes target samples of same categories cluster and samples of different categories separate so the two domains can be aligned on a finer-grained level.

\textbf{$\ell_p$-box ADMM} The binary constraint of $m_s$ brings extra difficulty. Fortunately, Wu \textit{ et al.} \cite{wu2018ell, li2019compressing} provide one algorithm called $\ell_p$-box ADMM to overcome such problem. It splits the constraint $m_s\in {\{ 0,1 \} }^N$ into two continuous constraint, $m_s \in S_b = {[0,1]}^N$ and $m_s \in S_p = {m: ||m - \frac{1}{2}||_2^2 = \frac{N}{4}}$. For any loss function $L$ (Eq. \ref{eq5} in this work) and sparsity threshold $t$, the optimization of $m_s$ can be formulated as:
\begin{equation}
    \underset{m_s,z_1,z_2}{argmin} \quad L(W, m_s) \quad s.t. \quad m_s = z_1 \quad m_s = z_2 \quad 1^{\top}z_1 = t \quad z_1 \in S_b \quad z_2 \in S_p
\end{equation}
We can extend the above formulation into the form of augmented Lagrangian function as:
\begin{equation}
\label{eq7}
     L(W,m_s) + h_1(z_1) + h(z_2) + u_1^{\top}(m_s - z_1) + u_2^{\top}(m_s - z_2) + \frac{\rho}{2}[||m_s - z_1||_2^2 + ||m_s - z_2||_2^2]
\end{equation}
Here, $h_1(z_1) = I[z_1 \in S_c: S_b \cap {z_1: 1^{T}z_1 = t}]$ and $h_2(z_2) = I[z_2 \in S_p]$. $u_1,u_2$ are the dual variables and $\rho$ is a penalty parameter. Then ADMM further splits Eq. \ref{eq7} into three independent sub-problems and $(m_s,z,u)$ will be alternately updated by each sub-problem.

\textbf{Update $\bf{m}$} $m_s$ can be updated by regular gradient descent as $m_s = m_s - \eta \frac{\partial L(W,m_s,z_1,z_2,u_1,u_2)}{\partial m_s}$. 

\textbf{Update $\bf{z_1}$, $\bf{z_2}$} They are updated by solving the following two problems respectively. 
\begin{equation} 
    z_1 = \text{argmin}_{z_1 \in S_c} \  \frac{1}{2}||z_1||_2^2 - z_1^{\top}(m_s + \frac{u_1}{\rho}) \quad z_2 = \text{argmin}_{z_2 \in S_p} \  \frac{1}{2}||z_2||_2^2 - z_2^{\top}(m_s + \frac{u_2}{\rho})
\end{equation}
Given the constraint of $S_c$, the first problem is a standard quadratic program (QP) problem. Then it can be solved by the mature existing solutions. The second problem can be solved by projecting the unconstrained problem onto $S_p$. 

\textbf{Update $\bf{u_1}$, $\bf{u_2}$} $u_1$ and $u_2$ are also updated with gradient descent. $u_1 = u_1 + z_1 - m_s$ and $u_2 = u_2 + z_2 - m_s$. 

In the experiment, we alternate the above two steps until the model converges. Although $m_h$ aims to worsen the model's performance on the target domain, we claim it can push the network to learn more robust $W$ and $m_s$. Besides, $m_h$ is dropped in fine-tuning. So it will not harm the final performance. We will provide the specific effect of $m_h$ on model performance to support the above claims in the ablation analysis part. 
\section{Experiment}
\subsection{Experimental settings}
\textbf{Dataset settings} We use two datasets Office31 and ImageCLEF-DA. These datasets are chosen because they are standard and maybe the most popular benchmarks for UDA. Office31 contains images of 31 categories from three different camera settings, Amazon(A), Webcam(W), and Dslr(D). We evaluate on all six transfer tasks of Office31: A $\rightarrow$ W, D $\rightarrow$ W, W $\rightarrow$ D, A $\rightarrow$ D, D $\rightarrow$ A, and W $\rightarrow$ A. ImageCLEF-DA collects images of 12 categories that are shared by three different public datasets, ImageNet ILSVRC 2012(I), Caltech-256(C) and Pascal VOC 2012(P). Like Office31, the six transfer tasks of ImageCLEF-DA are all evaluated in this work (I $\rightarrow$ P, P $\rightarrow$ I, I $\rightarrow$ C, C $\rightarrow$ I, C $\rightarrow$ P, and P $\rightarrow$ C). 

\textbf{Training settings} The experiments are conducted on ResNet \textit{e.g.}, ResNet50, as it is the mainstream backbone structure of UDA. In pretraining, we first use one regular UDA technique to train the model for 100 epochs. In pruning, we simply use a uniform pruning ratio to decide the pruning filters in each layer. We first use $lr=0.1$ to train the model about 1500 iterations so that the learning of $m_s$ becomes stable. Then the model will be trained about $5000$ iterations and the $lr$ is divided by 10 every $2000$ iteration. In fine-tuning, the model is trained about $6000$ iterations and early stop is used during fine-tuning.
\subsection{Comparisons with state-of-the-art
methods}
\label{compare}
In this section, we compare ADMP with TCP \cite{yu2019accelerating}, which is the current state of the art of cross-domain pruning. Table \ref{tab:office31} and Table \ref{tab:imageclef} have shown the comparison on Office31 and ImageCLEF-DA datasets respectively. The items in brackets mean relative accuracy loss. The minus sign means the accuracy after pruning is higher than that of the full-size baseline. Although we use the same way (DAN\cite{long2015learning}) to train the full-size baseline as in \cite{yu2019accelerating}, we find our baseline has similar accuracy with \cite{yu2019accelerating} on Office31 but much higher accuracy than \cite{yu2019accelerating} on ImageCLEF-DA. Hence, to be fair, the relative accuracy loss is calculated based on the respective baseline of \cite{yu2019accelerating} and ours. 

Results in Table \ref{tab:office31} and Table \ref{tab:imageclef} show the significant improvement of ADMP over the state of art. In general, our method can achieve less accuracy loss or even higher accuracy with higher pruning sparsity compared with previous works. On Office31, our method has an average $82.0\%$ accuracy when $38.1\%$ parameters are pruned. The pruned model even has a $0.5\%$ higher accuracy than the baseline. When $60.8\%$ parameters are pruned, the model still has $81.3\%$ accuracy. The corresponding accuracy loss is only $0.2\%$. While previous work only has $77.2\%$ accuracy with $36.7\%$ sparsity. Their accuracy loss reaches $4.3\%$. This means our ADMP can have at least $1.63\times$ parameters reduction under similar accuracy loss. Not to mention that we still have a $4.1\%$ accuracy improvement on Office31. Our advantages are more significant when pruning sparsity increases. For example, the accuracy loss of TCP reaches $10.2\%$ with $58.1\%$ sparsity while that of this work is only $0.2\%$. The case on ImageCLEF-DA is similar, so we will not go into the details here. 

It can be found that previous works perform well on easy transfer tasks but they degrade dramatically on the difficult ones. While our ADMP can keep good performance on all sub-datasets. On Office31, we can find tasks, such as D $\rightarrow$ W, W $\rightarrow$ D, are very easy so all methods can retain high accuracy after pruning. But if we investigate the difficult tasks like D $\rightarrow$ A, or W $\rightarrow$ A, ADMP is much more effective. On D $\rightarrow$ A and W $\rightarrow$ A, the accuracy loss of TCP reaches $13.6\%$ and $7.4\%$ even with $35.0\%$ and $36.9\%$ parameter reduction. But the proposed method only has $0.6\%$ and $-0.3\%$ accuracy loss even with a $60.8\%$ parameter reduction. The reason for the performance difference is that the teacher-student difference used in this work is a better measurement of domain discrepancy than MMD. Besides, the harder the task, the more samples may be distributed near the classification boundaries. The optimization direction of hard mask exacerbates this phenomenon. During adversarial training, the soft mask and weights are pushed to make up for the performance degradation by pushing samples away from the boundaries. So when the hard mask is dropped, the pruned model can have more robust decision boundaries.
\begin{table}[t]
    \centering
    \caption{Performance on Office-31 dataset (ResNet50-based). \textit{Acc} means the accuracy after pruning while the items in brackets mean relative accuracy loss. \textit{Param} $\downarrow$ means parameters reduction.}
    \scalebox{0.75}{
    \begin{tabular}{|c|c|c|c|c|c|c|c|c|}
    \toprule
    & \multicolumn{4}{|c|}{$\sim$40\% Param $\downarrow$} & \multicolumn{4}{|c|}{$\sim$60\% Param $\downarrow$} \\
    \hline
     Method & \multicolumn{2}{|c|}{TCP \cite{yu2019accelerating}} & \multicolumn{2}{|c|}{ADMP(This work)} & \multicolumn{2}{|c|}{TCP \cite{yu2019accelerating}} & \multicolumn{2}{|c|}{ADMP(This work)} \\
     \hline
    Sub-dataset & Acc  & Param $\downarrow$ & Acc & Param $\downarrow$ & Acc  & Param $\downarrow$ & Acc & Param $\downarrow$\\
    \hline
    A $\rightarrow$ W & 81.8\%(-1.5\%) & 37.7\% & \textbf{83.3\%}(-0.9\%) & \textbf{38.1\%} &77.4\%(2.9\%)& 58.4\%&\textbf{82.1\%(0.3 \%)}&\textbf{60.8\%}\\
    D $\rightarrow$ W & 98.2\%(-1.1\%) & 36.2\% & 98.9\%(-0.2\%) & \textbf{38.1\%} & 96.3\%(0.8\%) & 58.0\% &\textbf{98.6\%(0.1\%)} &\textbf{60.9\%}\\
    W $\rightarrow$ D & 99.8\%(-0.6\%) & 37.0\% & 99.9\%(-0.0\%) & \textbf{38.1\%} &100.0\%(-0.8\%)&57.1\%&99.9\%(-0.0\%) &\textbf{60.8\%}\\
    A $\rightarrow$ D & 77.9\%(1.0\%) & 36.9\% & \textbf{83.1\%(-1.4\%)} & \textbf{38.0\%} & 72.0\%(6.9\%) & 59.0\% &\textbf{81.5\%(0.2\%)} & \textbf{60.8\%} \\
    D $\rightarrow$ A & 50.0\%(14.3\%) & 35.0\% & \textbf{63.2\%(0.4\%)} & \textbf{38.1\%} &36.1\%(28.2\%)&57.8\%&\textbf{63.0\%(0.6\%)}&\textbf{60.9\%}\\
    W $\rightarrow$ A & 55.5\%(6.8\%) & 36.9\% & \textbf{64.2\%(-1.3\%)} & \textbf{38.1\%} & 46.3\%(16.0\%) & 58.5\% &\textbf{63.2\%(-0.3\%)} &\textbf{60.7\%}\\
    \hline
    Average & 77.2\%(3.1\%) & 36.7\% & \textbf{82.0\%(-0.5\%)} & \textbf{38.1\%} &
    71.3\%(9.0\%) & 58.1\% & \textbf{81.3\%(0.2\%)} &\textbf{60.8\%} \\
    \bottomrule
    \end{tabular}}
    \label{tab:office31}
\end{table}
\begin{table}[t]
    \centering
    \caption{Performance on ImageCLEF-DA dataset (ResNet50-based)}
    \scalebox{0.75}{
    \begin{tabular}{|c|c|c|c|c|c|c|c|c|}
    \toprule
    & \multicolumn{4}{|c|}{$\sim$40\% Param $\downarrow$} & \multicolumn{4}{|c|}{$\sim$60\% Param $\downarrow$} \\
    \hline
     Method & \multicolumn{2}{|c|}{TCP \cite{yu2019accelerating}} & \multicolumn{2}{|c|}{ADMP(This work)} & \multicolumn{2}{|c|}{TCP \cite{yu2019accelerating}} & \multicolumn{2}{|c|}{ADMP(This work)} \\
     \hline
    Sub-dataset & Acc  & Param $\downarrow$ & Acc & Param $\downarrow$ & Acc  & Param $\downarrow$ & Acc & Param $\downarrow$\\
    \hline
    I $\rightarrow$ P & 75.0\%(-0.2\%) & 37.5\% & \textbf{77.3\%}(-0.0\%) & \textbf{38.1\%} &67.8\% (7.0\%)& 57.2\%&\textbf{77.0\%(0.3 \%)}&\textbf{60.8\%}\\
    P $\rightarrow$ I & 82.6\%(-0.4\%) & 36.5\% & \textbf{90.2\%(-0.9\%)} & \textbf{38.1\%} & 77.5\%(4.7\%) & 58.0\% &\textbf{89.5\%(-0.2\%)} &\textbf{60.9\%}\\
    I $\rightarrow$ C & 92.5\%(-0.2\%) & 35.5\% & \textbf{95.8\%(-0.8\%)} & \textbf{38.1\%} &88.6\%(3.7\%)&56.2\%&\textbf{95.5\%(-0.5\%)} &\textbf{61.0\%}\\
    C $\rightarrow$ I & 80.8\%(2.5\% ) & 36.7\% & \textbf{88.9\%(0.8\%)} & \textbf{38.1\%} & 71.6\%(11.7\%) & 58.5\% &\textbf{88.9\%(0.8\%)} & \textbf{60.9\%} \\
    C $\rightarrow$ P & 66.2\% (3.8\%) & 36.6\% & \textbf{73.7\%(0.1\%)} & \textbf{38.1\%} &57.7\%(12.3\%)&55.7\%&\textbf{72.3\%(1.5\%)}&\textbf{61.0\%}\\
    P $\rightarrow$ C & 86.5\%(3.3\%) & 37.6\% & \textbf{91.8\%(-0.5\%)} & \textbf{38.1\%} & 79.5\%(10.3\%) & 58.2\% &\textbf{91.2\%(0.1\%)} &\textbf{60.7\%}\\
    \hline
    Average & 80.6\% (1.5\%) & 36.7\% & \textbf{86.3\%(-0.2\%)} & \textbf{38.1\%} &
    73.8\%(8.3\%) & 57.3\% & \textbf{85.7\%(0.3\%)} &\textbf{60.9\%} \\
    \bottomrule
    \end{tabular}}
    \label{tab:imageclef}
\end{table}
\subsection{Ablation analysis}
\subsubsection{Effect of $\ell_p$ ADMM}
Although the results in Table \ref{tab:office31} and Table \ref{tab:imageclef} have proved the advantages of the proposed ADMP over previous works. The reason for success should be further validated. ADMM is used in ADMP but many ADMM based works \cite{li2019compressing, ye2019progressive} have shown their advantages in pruning. Hence there is a possibility that the effectiveness of the proposed method only comes from the usage of ADMM. To answer such a question, we conduct the experiments shown in Fig. \ref{fig:admm_bar}. Apart from the mentioned methods in Section \ref{compare}, we provide one new control group termed ADMM + MMD. By investigating the details of TCP, we can find it uses a Taylor-based greedy algorithm to prune channels and an MMD-based loss to minimize domain discrepancy. Hence our new control group keeps the MMD-based loss but it uses ADMM to prune. The experiments are conducted on two difficult transfer tasks, $D \rightarrow A$ and $W \rightarrow A$ sub-datasets of Office31. Theses two sub-datasets are chosen because the effect of different methods can be more significantly shown and the impact of training randomness can be ignored on them. Fig. \ref{fig:admm_bar} shows the accuracy of ResNet50 when about $40\%$ parameters are pruned. It can be found that ADMM can improve the accuracy of the pruned model in some way while there is still a large gap between the control group \textit{ADMM + MMD} and ADMP. This shows the usage of ADMM cannot provide the whole improvement.
\subsubsection{Why adversarial}
\begin{figure}[tb]
\centering
\begin{minipage}[t]{0.45\textwidth}
\centering
\includegraphics[width=6.5cm]{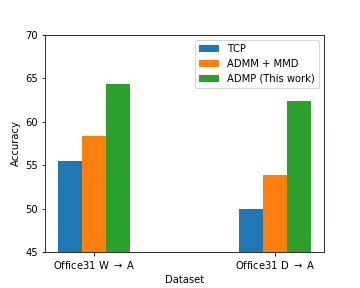}
\caption{(Best viewed in color) Effect of the introduction of ADMM}
\label{fig:admm_bar}
\end{minipage}
\begin{minipage}[t]{0.45\textwidth}
\centering
\includegraphics[width=6.5cm]{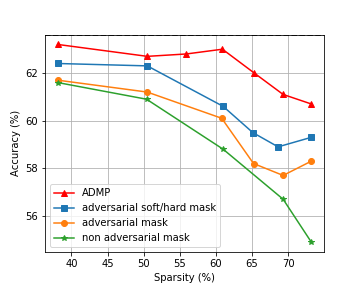}
\caption{(Best viewed in color) Effect of the different components in ADMP}
\label{fig:adv_plot}
\end{minipage}
\end{figure}

Another question that should be answered is how much effect does the proposed double adversarial masks have. It contains two sub-questions, what is the effect of introducing adversarial masks and why using double masks rather than one. There are two reasons that these two questions should be answered. First, when all the adversarial elements are removed, the framework degrades to a common KD framework. However, only using the KD framework can still relieve the problem of missing labels. So whether the adversarial pruning works, not just relying on the pseudo labels, needs to be tested. Second, we should know whether it is necessary to use two masks. In Section \ref{pruning}, it is mentioned that using two masks is due to the potential negative influence of $m_h$. The answers are illustrated in  Fig. \ref{fig:adv_plot}. We provide the accuracy-sparsity curves of ResNet50 under four different experimental settings on Office31 $D \rightarrow A$ because the impact of training randomness is smaller on this difficult transfer task. The red curve represents ADMP while the blue curve drops the clustering (SNTG) loss. The orange curve only uses one mask during pruning. That means it uses ADMM to learn an adversarial mask by minimizing Eq. \ref{eq4} while the weights are still updated by minimizing Eq. \ref{eq5} as that in ADMP. It can examine whether the hard mask has a negative influence. While the green curve removes the adversarial masks and just use ADMM to learn both the sub-structure and parameters by minimizing Eq. \ref{eq5}. In this way, it can be considered as a pure KD based pruning. 

The fact that the other three curves are above the green curve shows that methods with adversarial masks have higher accuracy under the same sparsity. The advantages are more obvious when the sparsity is high (The accuracy improvement is $5.8\%$, $4.4\%$, and $3.4\%$ respectively when sparsity is $73\%$). This shows the necessity of introducing adversarial masks and only using a KD framework to provide pseudo labels is not enough. On the other hand, the comparison between the blue and orange curves show the advantages of using two masks. Sometimes the improvement of the orange curve over the green curve is negligible and an unexpected great degradation even happens around $60\% - 70\%$ sparsity. The reason is that the adversarial exploration of one-mask pruning has the risk of ending at one bad searched structure. However, such problems can be solved with two masks. Given the fact that the adversarial hard masks are dropped before fine-tuning and only the soft masks decide the final structure, we can make sure the searched structure is always optimized for lowering target error. At the same time, the hard mask can still play the role to push the model parameters to find a more robust decision boundary and feature space. The red curve in Fig. \ref{fig:adv_plot} shows that ADMP has the highest accuracy under the same sparsity. This shows using the clustering loss can further reduce the negative influence of the hard mask and improve the effectiveness of the domain alignment. 
\section{Conclusion}
In this work, we propose a new pruning paradigm specialized for cross-domain compression that is termed ADMP. Based on a KD framework, two adversarial masks (soft and hard) are adopted in ADMP. On the one side, the double-mask structure effectively minimizes the domain discrepancy. On the other side, it also avoids ending at a sub-optimal pruned structure. Our comparison with the SOTA work has proven the effectiveness of this work and our advantages on difficult transfer tasks have shown that ADMP can help the pruned model have better decision boundaries that are away from the sample-dense area.   

\section{Broader Impact}
Nowadays, applying machine learning to end devices or even sensors, \textit{e.g.} TinyML, is attracting more and more attention. Both model compression and domain adaptation are very important for such applications because of the limited resources and data divergence. So the proposed cross-domain pruning method may be very helpful for this new trend and could promote deep learning to broader applications. Besides, our work can motivate pruning researchers to divert their attention to more scenarios. However, there are also some potential risks. The pruned model by the proposed method contains more information about the end data than traditional methods. If there is no strict protection of the compressed models and the end devices, user privacy may be stolen by attackers. Hence we recommend that any researcher using this work should strictly guarantee the safety of the model in actual use. 

{\small

\bibliographystyle {ieeetr} 
\bibliography{IEEEref}
}

\end{document}